# ISIC 2017 – Skin Lesion Segmentation Using Deep Encoder-Decoder Network

Ngoc-Quang Nguyen, Sang-Woong Lee
Department of IT Convergence Engineering, Department of Software, Gachon University, Seongnam,
South Korea ngocquang870@gmail.com, slee.gachon.ac.kr

*Abstract*—This paper summarizes our method and validation results for part 1 of the ISBI Challenge 2018. Our algorithm makes use of deep encoder-decoder network and novel skin lesion data augmentation to segment the challenge objective. Besides, we also propose an effective testing strategy by applying multi-model comparison.

## I. INTRODUCTION

Automatic skin lesion segmentation in dermoscopic images is a challenging task due to the low contrast between le-sion and the surrounding skin [1]. Besides, the irregular and fuzzy lesion borders, the existence of various artifacts, and various imaging acquisition conditions. In addition, artifacts and intrinsic cutaneous features, such as hairs, frames, blood vessels and air bubbles can make the automatic segmentation more challenging. However, this task is not trivial because melanoma usually has a large variety of appearance in size, shape, and color along with different types of skin and texture. In this paper, we present a fully automatic method for skin lesion segmentation by the encoder-decoder network that is trained continuously on prior knowledge of the data. For giving prediction, we combined multi-model and took the best results. Furthermore, we design a novel data augmentation for skin lesion segmentation due to the strong imbalance between the number of foreground and background pixels.

## II. LESION SEGMENTATION

### A. Dataset

We took part in the part I of ISBI Challenge 2017 - Skin Lesion Analysis Towards Melanoma Detection: Lesion Seg-mentation. The training dataset is given with 2000 dermoscopic images in .jpg format and the corresponding lesion labels in .png format. The images are given in many kind of sizes. The lesion types involved include nevus, seborrhoeic keratosis and malignant melanoma. The goal is to locate the position of various skin lesions against a variety of background and produce accurate binary masks for every images. Besides training set, the organizers provide a validation dataset that includes 100 images. The participants can submit the binary masks of these 150 images and evaluate the segmentation performance online. Additional test dataset with 1000 images is provided for final evaluation. The final rank is calculated based on Jaccard index which also known as IoU (Intersection over Union).

*Corresponding author

### B. Our Proposed Approach

*1) Data Augmentation:* The ability to generlize is a main research for the CNN. When the model is excessively hetero-geneous, for instance having too many paramiters compared to the number of training samples, overfitting can happend and weaken model's generalization ability. Thus the trained model might depict random error or noise in place of the underlying data distribution [3]. In unexpected cacses, the CNN model may have good performance on the training data, however it can fail dramatically in predicting prediod. To enhance the generalization ability of CNNs and to prevent overfitting, Not only we apply various data augmentation and regularization methods have been proposed, such as batch nor-malization [4], dropout [2], random cropping [5] and flipping [6].

Occlusion is one of critical influencing factor which can affect the performance of CNNs. In the other hand, when some parts of an objects are occluded, a good classification model could be able to recognize its category from the overall object aberration. However, the training dataset usually has limited variance in occlusion. In an ideal case when all the training objects are clearly visible and no occlusion, the trained model can deal with the testing images very well, due to the finite generalization ability of the model, may misclassify objects which are partially occluded. While, we can manually add the occluded images to traning data.

To improve the generalization ability of the model, this paper introduces the new data augmentation approach for skin lesion segmentation. Our approach is lightly similar with Dropout [2]. It prevents overfitting by discarding units of CNN with a propability p. But, 1) we operate on a continuous circle region, 2) no pixels are discarded and 3) we strongly focus on boosting to deal with occlusion and noise. By applying RCPV (Random hanging pixel value), we aim to enforce the model entirely learn about the edge of tumor rather than the whole part of tumor. The main goal of segmentaion is to locate the position of the object. Specially, in pursuit of colorectal segmentation, the important part of results are tumor's edges.

After making various expriments, we also realize that train-ing with graysale 3 chanels images could give better results. Hence, we turned all database into the form of the latter. Then, we apply most effective data augmentaton methods ([4], [5], [6]) to the training data. Not only we use these methods separately, we also combine them together to achieve better performance of the model in testing period.

```
Algorithm 1 Random Changing Pixel Value Algorithms
   Input : Input image I;
          Image size W and H;
          Area of tumor in ground truth dataset
          S; Changing probability p;
          Radius of circle $r_i$, $r_i$ Rand (0,R) with R is
          distance from tumor center $C(x_C, y_C)$ to
          closest edge's pixel;
   Output: Processed image I
   Initialization: $p_1$ ← Rand (0,1).
1  if $p_1$ ≤ p then
2       I ← I;
3       return I .
4  else
5       While True do
6           $r_i$ ← Rand (0,R);
7           $S_e$ ← $r_i^2$;
8           $x_e$ ← Rand ($x_C$, $x_C$ + R);
9           $y_e$ ← Rand ($y_C$, $y_C$ + R);
10          if $x_e$ ≤ $x_C$ ≤ R and $y_e$ ≤ $y_C$ ≤ R then
11              $I_e$ ← ($x_C, y_C, r_i$);
12              $I(I_e)$ ← Rand(0,128);
13              I ← I;
14              return I .
15          end
16      end
17 end
```

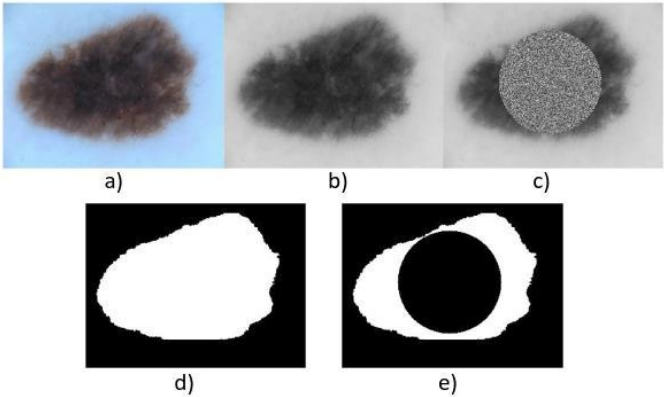

Fig. 1: Proposed data augmentation for skin lesion segmentation a) RGB image, b) gray scale image, c) processed image, d) label image, e) processed label image

2) Architecture: Spatial pyramid pooling module or encode-decoder structure are used in deep neural networks for semantic segmentation task. The former networks are able to encode both single-scale and multi-scale contextual information and by probing the incoming features with filters or pooling operations at multiple rates and multiple effective fields-of-view, while the latter networks can capture sharper object boundaries by gradually recovering the spatial information [7]. For producing prediction image we did resize image to the original dimension of testing image. Besides, we applied proposed image augmentation and other method as mention before to further improve the robustness of the proposed model under a wide variety of image acquisition conditions. Note that these augmentations only require little extra computation, so the transformed images are generated from the original images for every mini-batch within each iteration.

Not we apply one of the most powerful segmentation architecture for training, but we also combine 3 trained models to robustly predict the segmetation masks. In the first model, the network operated on an input grayscale image of 500 350 pixels, when the second former was trained with 450 300 pixel. Mean while, we gave the 400 250 pixels images as the inputs of the network. After comparing the results which are produced by each of models, we selected best prediction that have the highest Jaccard index.

3) Post-processing: Due to the dimension of testing images, we use changing image size two times: one for input period and another one for turning the prediction image backs to integrant proportion. Before submiting to the validation stage, we used morphological transformation algorithms to make small holes or noises disappeard. The kernel slides through the image (as in 2D convolution). A pixel in the original image (either 1 or 0) will be considered 1 only if all the pixels under the kernel is 1, otherwise it is eroded (made to zero). Here, we applied 10 10 kernel with full of ones.

### III. EXPERIMENTS AND RESULTS

Our proposed method was implemented with Python and Tensorflow. The experiments were conducted on a Intel i7-7700 and a GPU of Nvidia GeForce GTX 1080Ti with 8GB DDR5 memory.

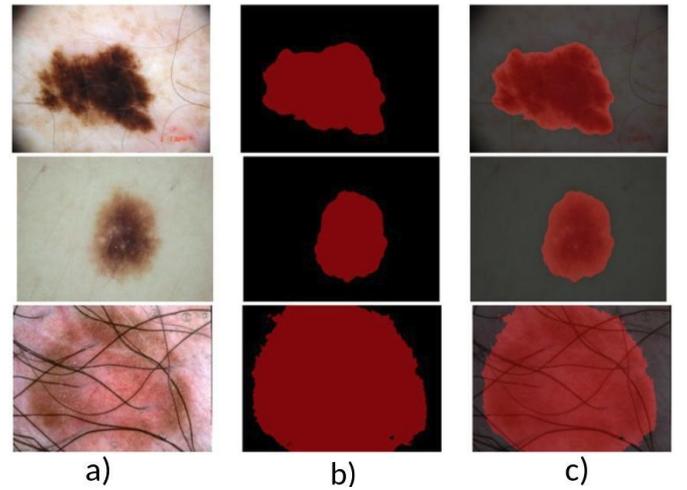

Fig. 2: Results from proposed approach a) RGB images, b) prediction results image, c) segmentation overlay images.

Finally, our method achieved an average Jaccard index of 0.808 on the online validation dataset and our proposed approach has shown the better performance than the first candidate in 2017 challenge.

3## REFERENCES

[1] Y. Yuan et al., "Automatic skin lesion segmentation using deep fully convolutional networks with jaccard distance," in IEEE Trans. Med. Imaging (2017), 10.1109/tmi.2017.2695227.

[2] N. Srivastava et al., "Dropout: a simple way to prevent neural networks from overfitting," in Journal of Machine Learning Research, 1929-1958 (2014).

[3] C. Zang et al., "Understanding deep learning requires rethinking generalization," in International Conference on Learning Representations, 2017.

[4] S. Ioffe and C. Szegedy, "Batch nomalization: Accelerating deep network training by reducing internal covariate shift," in International Conference on Machine Learning, 2015.

[5] A. Krizhevsky et al., "Understanding deep learning requires rethinking generalization," in International Conference on Learning Representa-tions, 2017.

[6] K. Simonyan, A. Zisserman, "Very deep convolutional networks for large-scale image recognition" in International Conference on Learning Representations, 2015.

[7] C Liang-Chieh et al., "Encoder-Decoder with Atrous Separable Convolution for Semantic Image Segmentation". arXiv:1802.02611, 2018.